\documentclass[conference, letterpaper]{IEEEtran}
\usepackage{soul}
\IEEEoverridecommandlockouts

\usepackage[pdftex]{graphicx}  

\usepackage{multirow}
\usepackage{tikz}
\usetikzlibrary{arrows.meta, positioning}
\usepackage{graphicx}
\usepackage{subcaption}
\usepackage{booktabs}
\usepackage[left=0.625in, top=0.75in, right=0.625in, bottom=1.35in]{geometry}
\usepackage{amsmath,amsfonts}
\usepackage{algorithm,algorithmic}
\usepackage{url}

\begin{document}


\title{Precise Top-Layer Fabric Segmentation \\ for Fabric Destacking \\ with Edge- and Shape-Aware Deep Networks}

\author{
\IEEEauthorblockN{
    Wenbo Dong\textsuperscript{1,2},
    Dipankar Bhattacharya\textsuperscript{1,2},
    Akinari Kobayashi\textsuperscript{1,2},
    Akira Seino\textsuperscript{1,2},
    Fuyuki Tokuda\textsuperscript{1,2},\\
    Xuzhao Huang\textsuperscript{1,2},
    Kai Tang\textsuperscript{1,2},
    Norman C. Tien\textsuperscript{2,3},
    and Kazuhiro Kosuge\textsuperscript{1,2}
    }\\
\IEEEauthorblockA{
\textit{\textsuperscript{1}JC STEM Lab of Robotics for Soft Materials, Department of Electrical and Electronic Engineering,}\\
\textit{Faculty of Engineering, The University of Hong Kong, Hong Kong SAR, China}\\
\textit{\textsuperscript{2}Centre for Transformative Garment Production, Hong Kong SAR, China}\\
\textit{\textsuperscript{3}Department of Electrical and Electronic Engineering,}\\
\textit{Faculty of Engineering, The University of Hong Kong, Hong Kong SAR, China}\\
\textit{\textsuperscript{*}Corresponding Author: dongwbo@connect.hku.hk}\\
\textit{Code: \url{https://github.com/bhattner143/top-layer-fab-seg}}
}
}

\makeatletter
\setlength{\footskip}{18pt}
\def\ps@IEEEtitlepagestyle{%
  \def\@oddhead{\@empty}%
  \def\@evenhead{\@empty}%
  \def\@oddfoot{%
    \raisebox{-\height}{\parbox[t]{\columnwidth}{\scriptsize\raggedright
    \copyright~2025 IEEE. Personal use of this material is permitted.
    Permission from IEEE must be obtained for all other uses, in any current or future media,
    including reprinting/republishing this material for advertising or promotional purposes,
    creating new collective works, for resale or redistribution to servers or lists,
    or reuse of any copyrighted component of this work in other works.\\
    Published in: 2025 IEEE International Conference on Mechatronics and Automation (ICMA),
    Beijing, China, 3--6 Aug.\ 2025,
    doi: 10.1109/ICMA65362.2025.11120697.}}%
    \hfill}%
  \def\@evenfoot{\@empty}%
}
\makeatother

\maketitle

\begin{abstract}
Fabric destacking requires precise segmentation of the topmost fabric layer, a task complicated by subtle fabric boundaries and high visual similarity between fabric layers.
Existing semantic and edge-based segmentation approaches often struggle with these complexities, limiting the performance of robotic manipulation for different tasks.
In this work, a novel segmentation training architecture tailored for top-layer fabric segmentation in stacked fabrics is proposed. The method extends the classical encoder-decoder framework by introducing two specialized branches—an edge-aware branch and a shape-aware branch—that are used to supervise the backbone network for better tuning. The edge-aware branch enhances boundary delineation, while the shape-aware branch guides the network to capture and align the overall fabric shape with reference masks derived from Computer Aided Design (CAD) models.
Experiments on a real-world fabric dataset demonstrate that the training approach outperforms established baselines, verifying the effectiveness of the multi-branch design through both quantitative results and ablation studies. 
\end{abstract}


\section{Introduction}

\begin{figure}[t]
    \centering
    \begin{subfigure}[b]{0.7\linewidth}
        \centering
        \includegraphics[width=\linewidth]{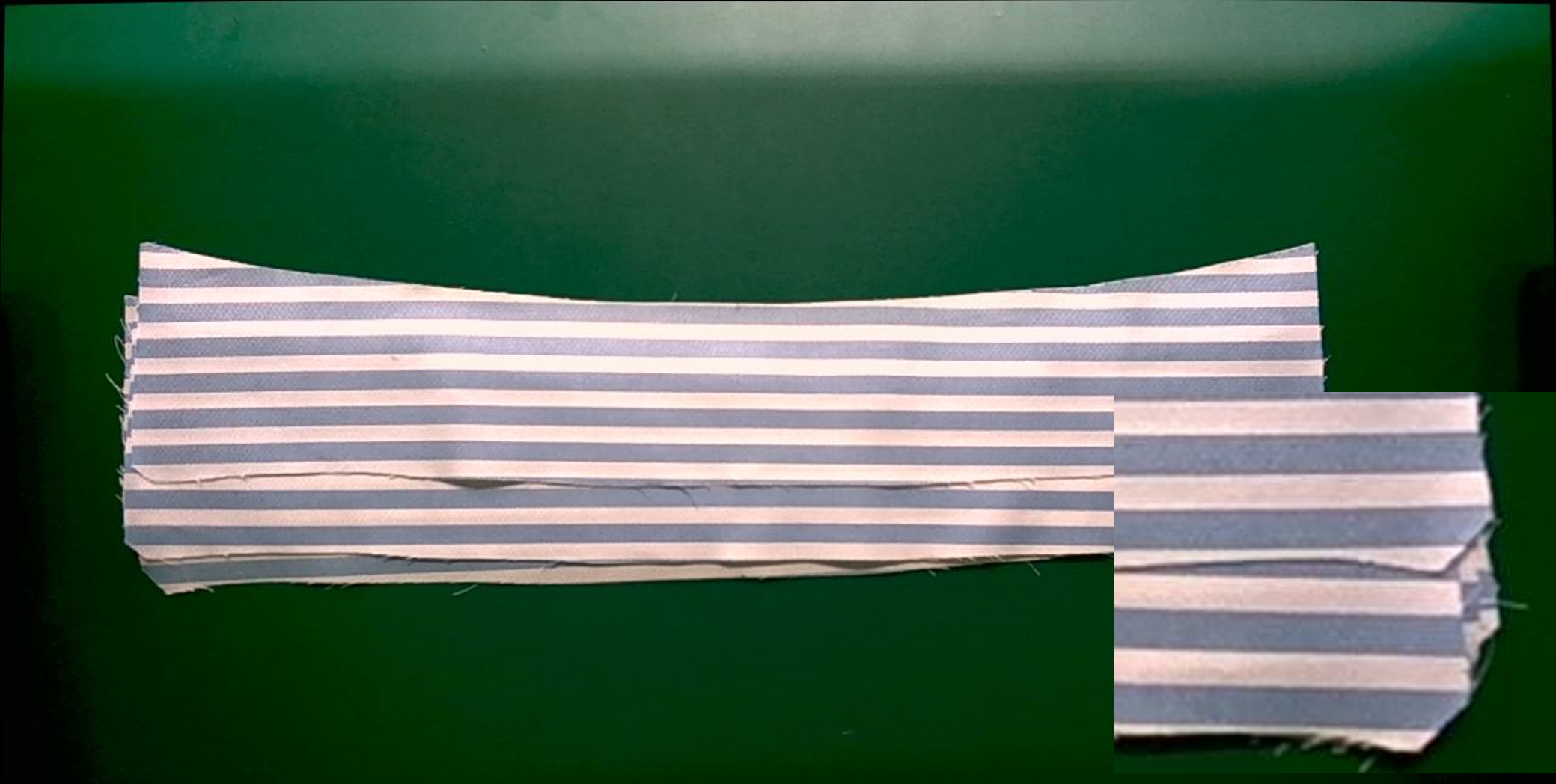}
        \caption{Stacked fabric layer image.}
        \label{fig:org}
    \end{subfigure}
    \\
    \begin{subfigure}[b]{0.7\linewidth}
        \centering
        \includegraphics[width=\linewidth]{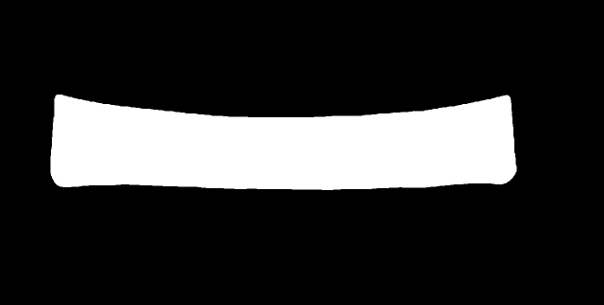}
        \caption{Top-layer segmentation.}
        \label{fig:result}
    \end{subfigure}
    \caption{Fabric stack image and top-layer segmentation.}
    \label{fig:vis-example}
\end{figure}

Fabric destacking is a complex process that increasingly relies on automation to enhance efficiency and consistency in garment production. A critical and challenging aspect of this automation is the manipulation of the top-layer fabric from fabric stacks (Fig.~\ref{fig:vis-example} (a)), which underpins key manufacturing steps such as picking and placing~\cite{seino2025passive}, fabric alignment, folding, and sewing of flexible fabric components~\cite{tang2024time, tokuda2024fixture}. 
Therefore, generating accurate \textit{top-layer segmentation} (Fig.~\ref{fig:vis-example} (b)) is essential for enabling reliable grasping, unfolding, and sewing automation.
However, achieving precise top-layer segmentation in real-world settings presents several unique challenges:
(1) the boundaries between stacked fabric layers are often subtle, irregular, or visually ambiguous; (2) the top layer and the background typically exhibit highly similar color and texture features, as similar fabrics are commonly stored together; and (3) imperfections along the boundaries of real fabrics can further interfere with the accurate segmentation of the topmost layer.

Conventional \textit{semantic segmentation methods}, which classify each pixel in an image of clothing into different parts or types of garments (such as sleeves, collars, or skirts), have achieved significant progress in generic image segmentation tasks. Architectures like UNet~\cite{ronneberger2015u} are notable examples of this advancement. 
However, such methods do not effectively handle subtle boundaries and high similarity between stacked fabric layers and the background.
To address this, \textit{edge-based segmentation methods} approaches such as GSCNN~\cite{takikawa2019gated}, have proven effective in scenarios where object contours are ambiguous or subtle. 
However, these methods often struggle when faced with the high similarity and complex boundaries characteristic of layered fabric scenarios.
\begin{figure*}[!th]
    \centering
    \includegraphics[width=0.9\textwidth]{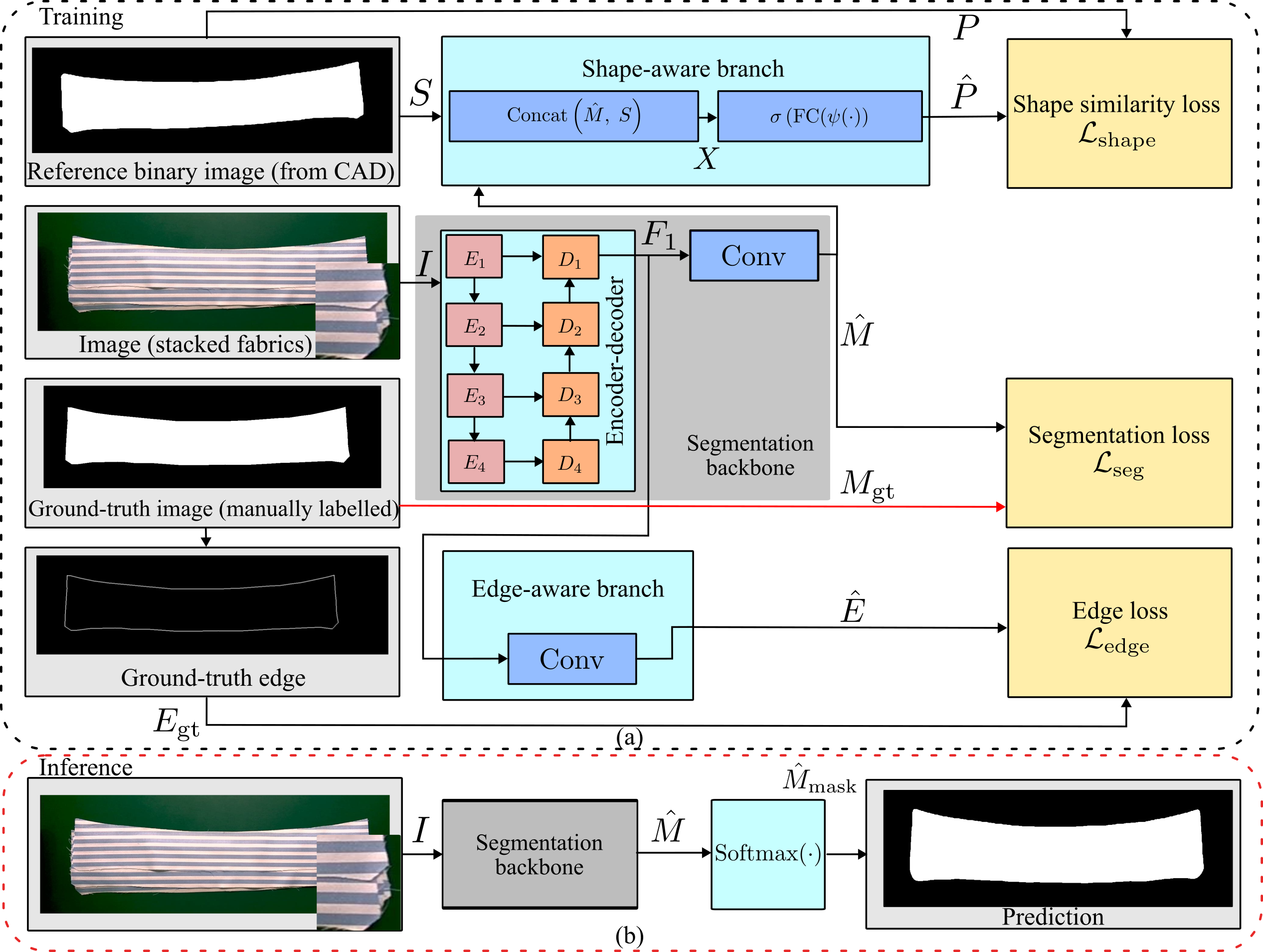}
    \caption{Proposed training architecture and inference scheme.}
    \label{fig:network-arch}
\end{figure*}
In summary, current approaches often fall short in enabling precise and efficient top-layer fabric segmentation, directly affecting the performance of robotic manipulation. 

Hence, there remains a clear need for a robust method that can accurately handle the fabric's subtle boundaries and high visual similarity while remaining practical for real-world use. 

To address these challenges in segmenting the top-layer in stacked fabrics, this work proposes a novel training architecture, as shown in Fig.~\ref{fig:network-arch} (a), consisting of two additional branches to enhance the segmentation accuracy during the training of the segmentation backbone network. The architecture builds upon the classical \textit{encoder-decoder architecture} by incorporating two specialized branches for training supervision.
\begin{itemize}
    \item \textbf{{Edge-aware branch}:} The edge-aware branch is specifically added to focus on the edges of the fabric. This helps supervising the segmentation backbone network to accurately outline the edges of the top-layer fabric and overcome the problem of unclear or blurry boundaries that often occur with standard segmentation methods.
%
    \item \textbf{Shape-aware branch:}  The shape-aware branch focuses on understanding the overall shape of the fabric by learning features that represent the entire predicted fabric segmentation region. It aligns these features with an ideal reference binary image from the CAD model, serving as a regularization step to ensure segmentation results are accurate both at the pixel level and in overall shape.
\end{itemize}

As a result of adding these branches, the trained backbone network can infer segmented top-layer fabric (Fig. \ref{fig:network-arch} (b)) that remains realistic and consistent with actual fabric shapes, even when the input image contains noise or visually ambiguous areas. It should be noted that for inference, only the segmentation backbone is used, with weights enhanced by edge and shape-aware supervision for improved results.
Experiments conducted on a real-world fabric dataset demonstrate that the method achieves superior accuracy compared to other models. Ablation studies further confirm the increase of the segmentation accuracy by adding the two branches during training. Fig.~\ref{fig:vis-example} (a) shows an example input image of stacked fabrics, while Fig.~\ref{fig:vis-example} (b) provides the segmentation result obtained using the proposed method.

The remainder of the paper is organized as follows. The related work is reviewed in Section II. Section III provides the methodology to design the training architecture.
Section IV provides the
experimental results, which includes ablation studies and qualitative results, and finally Section V provides the conclusion.

\section{Related Work}

\textbf{Semantic segmentation methods:} Encoder-decoder architectures, such as UNet~\cite{ronneberger2015u}, SegNet~\cite{badrinarayanan2017segnet}, and the DeepLab series~\cite{chen2018encoder}, have become the standard paradigm for semantic segmentation tasks. These models extract hierarchical features via the encoder and progressively restore spatial resolution through the decoder, often employing skip connections to recover fine details. Recent advances, including atrous convolutions, pyramid pooling~\cite{zhao2017pyramid}, and attention mechanisms~\cite{fu2019dual}, have further improved segmentation accuracy by enhancing multi-scale context and feature representation. However, these general-purpose frameworks do not explicitly address the subtle boundaries and high similarity between stacked fabric layers and the background, which are critical challenges in layered fabric segmentation.

\textbf{Edge-based segmentation methods:}
In scenarios where object contours are ambiguous or subtle, precise edge localization is essential for achieving high-quality segmentation. GSCNN~\cite{takikawa2019gated} proposes a Gated-Shape Convolutional Neural Network (CNN) architecture that fuses edge and region cues via a gated module. In contrast, DCAN~\cite{chen2016dcan} introduces deep contour-aware supervision to produce sharper mask boundaries. CASENet~\cite{yu2017casenet} and BiseNet~\cite{yu2018bisenet} leverage edge-aware branches or edge-sensitive modules to improve contour prediction, and BASNet~\cite{qin2019basnet} as well as EGNet~\cite{zhao2019egnet} further focus on edge refinement for salient object segmentation. However, these methods work for generic object or segmenting objects which are visually prominent, and do not sufficiently address the highly similarity characteristic of layered fabric scenarios. 
In contrast, the proposed training approach integrates a lightweight edge detection branch supervising the segmentation backbone, providing direct edge supervision for the unique challenges of top-layer fabric segmentation, thereby further enhancing accuracy and robustness.

\textbf{Shape-based segmentation methods:}
Incorporating global shape information has proven effective for regularizing segmentation outputs, especially in domains requiring anatomical or geometric consistency. Some methods employ adversarial training or shape priors to encourage plausible segmentation boundaries~\cite{oktay2017anatomically, xue2018segan}, while others utilize distance transforms or shape-aware losses as auxiliary supervision to enforce geometric regularity~\cite{chen2024ld}. However, many of these approaches are designed for anatomical or organ segmentation tasks where the target and background are relatively well-defined and distinct. They often require additional annotation or complex training strategies, and may introduce significant computational overhead. 
In contrast, the proposed method adopts a compact CNN-based shape feature extractor that directly aligns the predicted segmentation with ideal shape features, coming from the CAD model. This design not only reduces computational complexity but also provides effective global regularization on the fabric shape, making it well-suited for the unique challenges of top-layer fabric segmentation.

\textbf{Multi-task and auxiliary supervision methods:}
Multi-task learning has been widely used to improve segmentation by leveraging auxiliary tasks such as edge detection, depth estimation, or object detection~\cite{chen2016dcan,kendall2018multi, kokkinos2017ubernet, zhang2019net, xie2022end, caruana1997multitask}. Jointly optimizing multiple objectives enables the network to learn richer representations and improves generalization, particularly when labeled data is limited. Previous works have explored multi-branch or unified architectures that integrate auxiliary cues to enhance segmentation accuracy~\cite{chen2016dcan,kokkinos2017ubernet,zhang2019net}. However, many of these approaches focus on generic object scenes with clear boundaries.
In contrast, the proposed method is designed as a lightweight, task-specific multi-branch architecture, where each branch (fabric edge and shape) addresses a distinct structural aspect of the fabric relevant to layered fabric segmentation. 

\section{Methodology}
To address the challenge of segmenting the topmost fabric layer, this section first discusses the task definition and then presents the proposed novel training architecture. The architecture, illustrated in Fig.~\ref{fig:network-arch}, builds upon a classical encoder-decoder segmentation backbone and introduces additional edge-aware and shape-aware branches for training the backbone.


\subsection{Task definition and supervised annotations}

Let $I \in \mathbb{R}^{H \times W \times 3}$ denote the input RGB image of a fabric stack (Fig~\ref{fig:network-arch}), where $H$ and $W$ represent the image height and width, and $3$ corresponds to the color channels (red, green, and blue). 
The aim is to train a segmentation network that predicts a \textit{segmentation probability map} $\hat{M} \in [0, 1]^{H \times W}$, where each element $\hat{M}_{i,j}$ indicates the probability that pixel $(i, j)$ belongs to the top-layer fabric. A binary segmentation mask can then be obtained by 
\begin{equation}
    \hat{M}_{\text{mask}}=\mathrm{Softmax}(\hat{M}),
\end{equation}
where \(\mathrm{Softmax}\) denotes softmax function and \(\hat{M}_{\text{mask}}\in\{0,1\}^{H\times W}\). This process is illustrated in Fig.~\ref{fig:network-arch} (b).
%
%
Hence, to obtain such a trained network and provide rich structural supervision, each training sample is annotated with the following
\begin{itemize}
    \item \textbf{Segmentation ground-truth mask} $M_{\text{gt}} \in \{0,1\}^{H \times W }$: Manually labeled mask indicating the top-layer fabric region.
    \item \textbf{Ground-truth edge mask} $E_{\text{gt}} \in \{0,1\}^{H \times W}$: Derived from $M_{\text{gt}}$ to highlight the top-level fabric edges.
    \item \textbf{Reference shape mask} $S \in \{0,1\}^{H \times W}$: A binary mask generated from the CAD model, representing the reference shape of the top-layer fabric in the image plane.
    

    \item \textbf{Shape alignment label} $P \in \{0,1\}$: A binary label indicator showing whether the predicted segmentation mask is aligned with the reference shape ($P=1$ for alignment, $P=0$ otherwise).
\end{itemize}
During training, the network receives the image $I$ as input and is jointly supervised by these annotations, enabling accurate segmentation, precise boundary localization, and keep the overall fabric shape consistent.

\subsection{Training network architecture overview}

As depicted in Fig.~\ref{fig:network-arch} (a), the proposed training architecture consists of three main components: a segmentation backbone, an edge-aware branch and a shape-aware branch.

\subsubsection{Segmentation backbone} 
The network adopts an {encoder-decoder structure with skip connections}.
The encoder utilize a \textit{pre-trained ResNet50} backbone to extract hierarchical features from the input image at multiple spatial resolutions. 
As the image is processed through successive layers of the encoder, the spatial dimensions are progressively reduced while the depth of the feature representations increases. 
This process produces a set of feature maps, denoted as $\{F_1, F_2, F_3, F_4\}$, where $F_1$ corresponds to \textit{low-level, high-resolution features}, and $F_4$ captures high-level, low-resolution semantic information. 
Each feature map $F_i$ encodes information at a different level of abstraction, enabling the network to learn both fine-grained details and global context. 

The decoder reconstructs the segmentation mask by progressively upsampling these encoded features through a sequence of upsampling modules. At each stage, the decoder concatenates the upsampled output from the previous layer with the corresponding high-resolution feature map from the encoder via skip connections. This fusion preserves fine-grained spatial details that are critical for accurate segmentation. Each concatenated feature map is then refined by a stack of convolutional layers and nonlinear activations.

In this architecture, the decoder output $F_1$ from the final upsampling block $D_1$ is used for generating the segmentation mask. This choice is made because $F_1$ has the highest spatial resolution among all decoder outputs, matching the input image size. Utilizing $F_1$ ensures that the predicted segmentation mask preserves fine-grained spatial details and aligns accurately with the original input. The predicted segmentation probability map is obtained by applying  
\begin{equation}
    \hat{M} = \sigma\left(\mathrm{Conv}(F_1)\right),
\end{equation}
where \(\mathrm{Conv}(\cdot)\) is convolution and \(\sigma(\cdot)\) is sigmoid activation function.


\subsubsection{Edge-aware branch}

To explicitly address the challenge of imprecise object boundaries, an edge-aware branch is added to the proposed network architecture, operating in parallel with the segmentation backbone. This branch takes \(F_1\) as input and processes it through a convolutional layers followed by nonlinear activations. The resulting feature map is then projected to a single-channel edge logits map via a $1\times1$ convolution. A sigmoid activation is then applied to obtain a \textit{predicted edge probability map} $\hat{E} \in [0,1]^{1 \times H \times W}$, given by
\begin{equation}
    \hat{E} = \sigma\left(\mathrm{Conv}(\phi(F_{1}))\right),
\end{equation}
 where each value indicates the likelihood of a pixel belonging to an object edge. Here, $\phi(\cdot)$ denotes the sequence of convolutional layers with ReLU activations.

Supervision for the edge-aware branch is provided by the ground-truth edge mask $E_{\text{gt}}$, encouraging the network to capture fine-grained and accurate object contours. This explicit edge modeling guides the network to better distinguish between adjacent objects and background, leading to sharper and more precise segmentation results.

\subsubsection{Shape-aware branch}

To regularize the global structure of the predicted fabric masks and suppress false positives, a shape-aware branch is implemented as a lightweight CNN. This branch begins by constructing a two-channel tensor, $X$, formed by concatenating the predicted segmentation probability map \(\hat{M}\) and the reference shape label \(S\in \{0,1\}^{H \times W}\) generated from the CAD model. The concatenated tensor can be written as
\begin{equation}
    X = \mathrm{Concat}\left(\hat{M},\; S\right) \in \mathbb{R}^{2 \times H \times W},
\end{equation}
Next, the input $X$ is processed by a CNN, denoted by $\psi(\cdot)$, followed by a fully connected layer \(\mathrm{FC}\) and a sigmoid activation to produce a predicted scalar probability
\begin{equation}
    \hat{P}= \sigma\left(\mathrm{FC}(\psi(X_{\text{}}))\right),
\end{equation}
where $\hat{P} \in [0,1]$ indicates the predicted likelihood that the segmentation aligns with the ideal shape.

\textbf{Pre-training:} To provide a robust understanding of the fabric shape, the shape-aware branch is first pre-trained as an independent binary classifier on synthetic mask pairs. The pre-trained weights are then used to initialize the shape-aware branch, allowing it to be further fine-tuned jointly with the entire network. 
This approach gives the network effective global structural guidance, helping it distinguish between plausible from implausible shapes, and improving both segmentation reliability and edge accuracy.

\subsubsection{Training losses}

The network is trained with a composite loss (Fig.~\ref{fig:network-arch} (a), yellow blocks) that jointly optimizes segmentation accuracy, \textit{edge and shape regularity}. The overall loss function is defined as
\begin{equation}
    \mathcal{L}_{\text{total}} = \mathcal{L}_{\text{seg}} 
    + \lambda_{\text{edge}}\, \mathcal{L}_{\text{edge}} 
    + \lambda_{\text{shape}}\, \mathcal{L}_{\text{shape}},
\end{equation}
where $\lambda_{\text{edge}}$ and $\lambda_{\text{shape}}$ are hyperparameters that balance the contributions of the edge and shape-aware branches.
The segmentation loss $\mathcal{L}_{\text{seg}}$ combines two components: the standard pixel-wise cross-entropy loss, which checks how well each pixel in the predicted mask matches the ground-truth mask, and the Dice loss, which measures the overall similarity between the predicted segmentation mask $\hat{M}$ and the ground-truth mask $M_{\text{gt}}$.
The edge loss $\mathcal{L}_{\text{edge}}$ is defined as the binary cross-entropy between the predicted edge mask \(\hat{E}\) and the ground-truth edge mask $E_{\text{gt}}$, thereby promoting precise boundary detection. For the shape-aware branch, $\mathcal{L}_{\text{shape}}$ is computed as the binary cross-entropy between the predicted scalar probability $ \hat{P}$ and the shape label $P$, which indicates whether the predicted segmentation matches with the ideal shape.


\subsubsection{Inference}
During inference (Fig. \ref{fig:network-arch} (b)), only the segmentation backbone network is used taking a camera image as input and producing a segmented mask as output. Additionally, the network's weights have been improved during training with supervision from the edge and shape-aware branches to achieve better segmentation results. The final segmentation output is produced using softmax and argmax operations. 
\section{Experiments}

\begin{figure}[t]
    \centering
    \setlength{\tabcolsep}{1pt} 
    \renewcommand{\arraystretch}{0.98} 
    \begin{minipage}[t]{0.154\textwidth}
        \centering
        \includegraphics[width=0.95\textwidth]{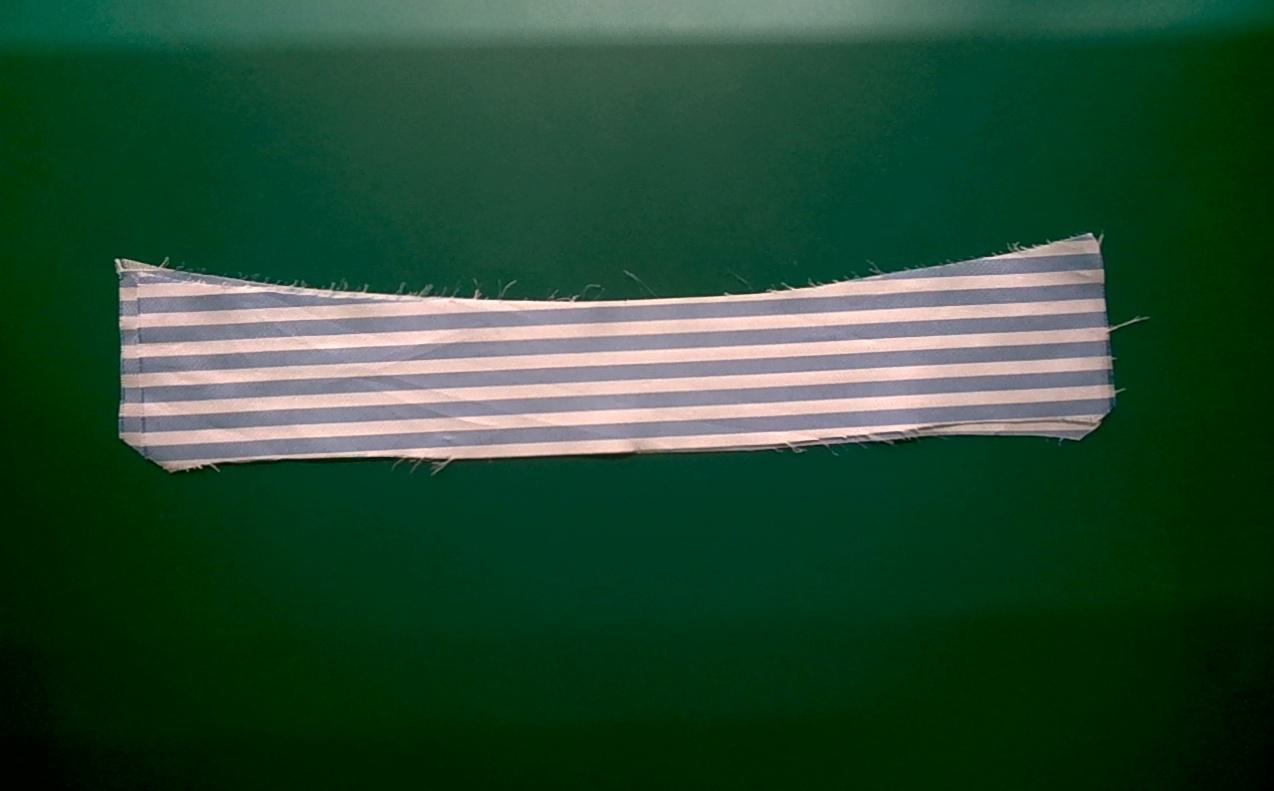}\par\vspace{0.07\textwidth}
        \includegraphics[width=0.95\textwidth]{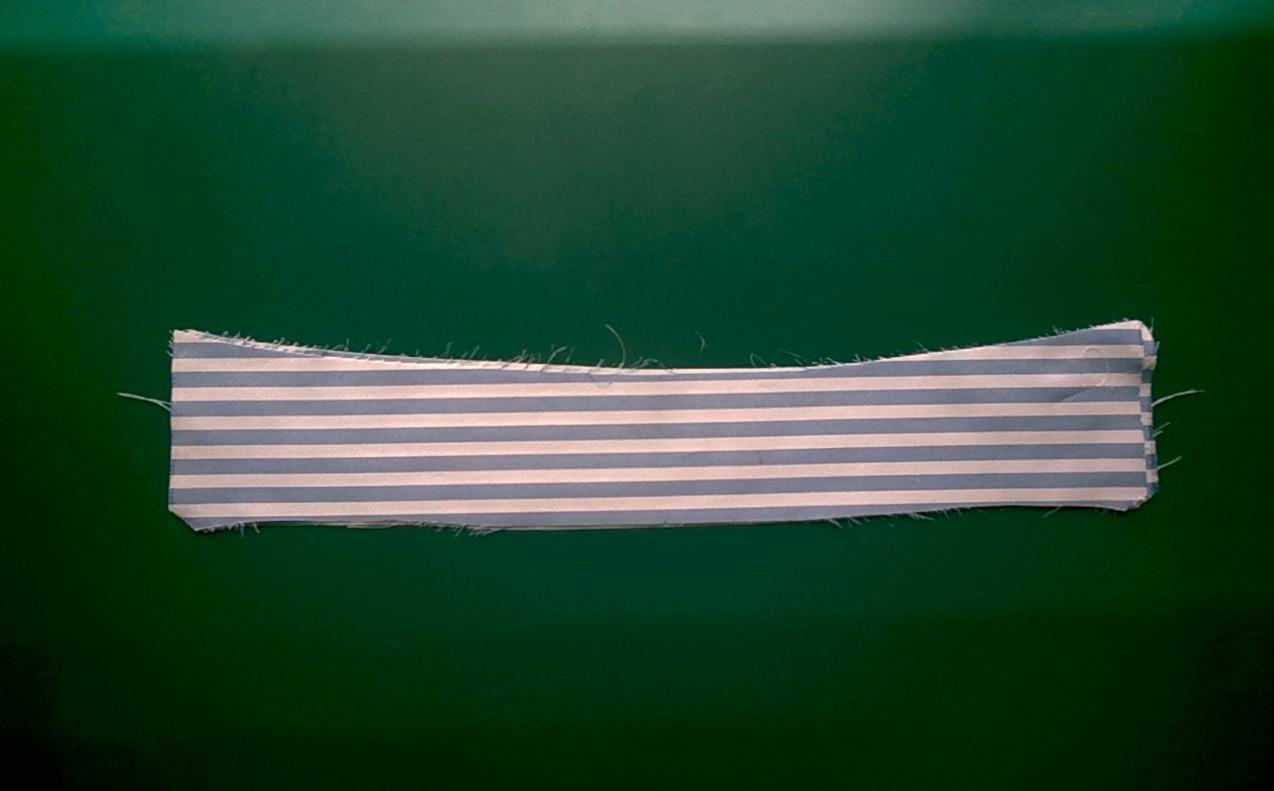}\par\vspace{0.07\textwidth}
        \includegraphics[width=0.95\textwidth]{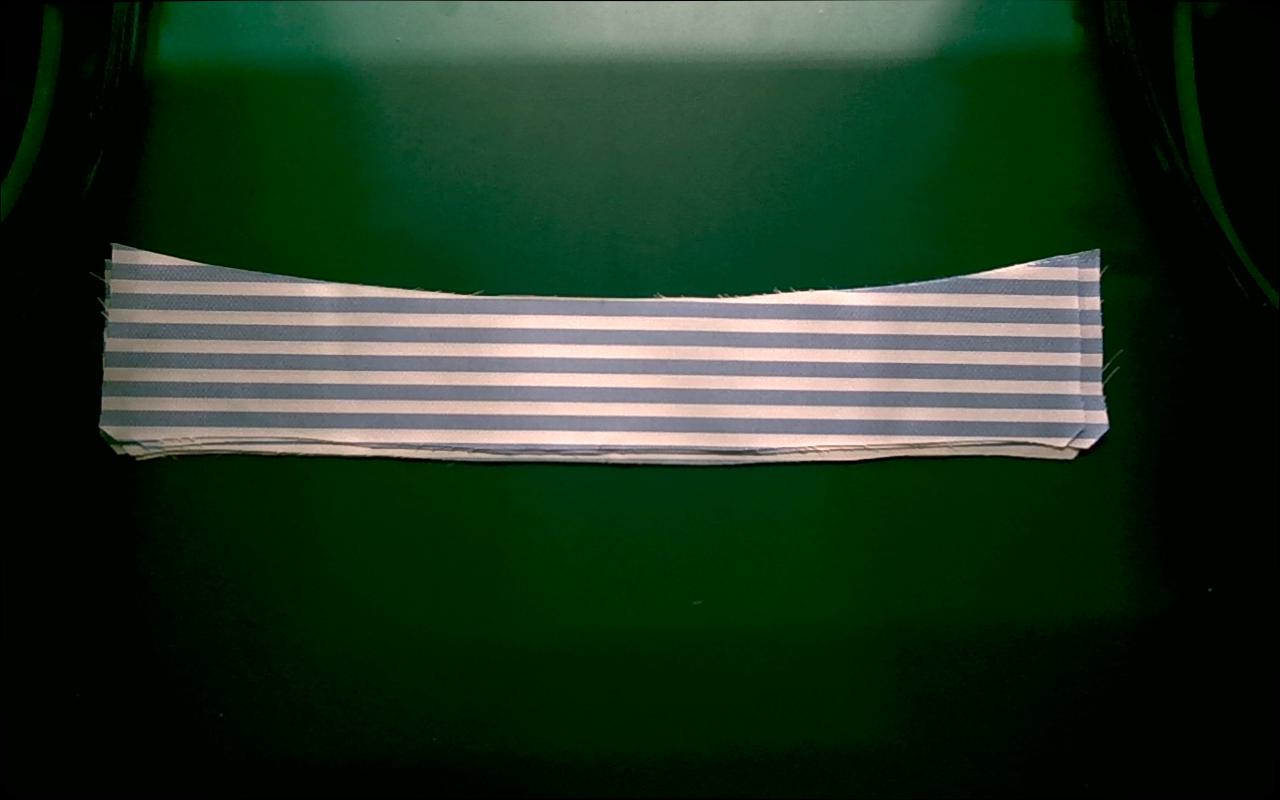}\par\vspace{0.07\textwidth}
        \includegraphics[width=0.95\textwidth]{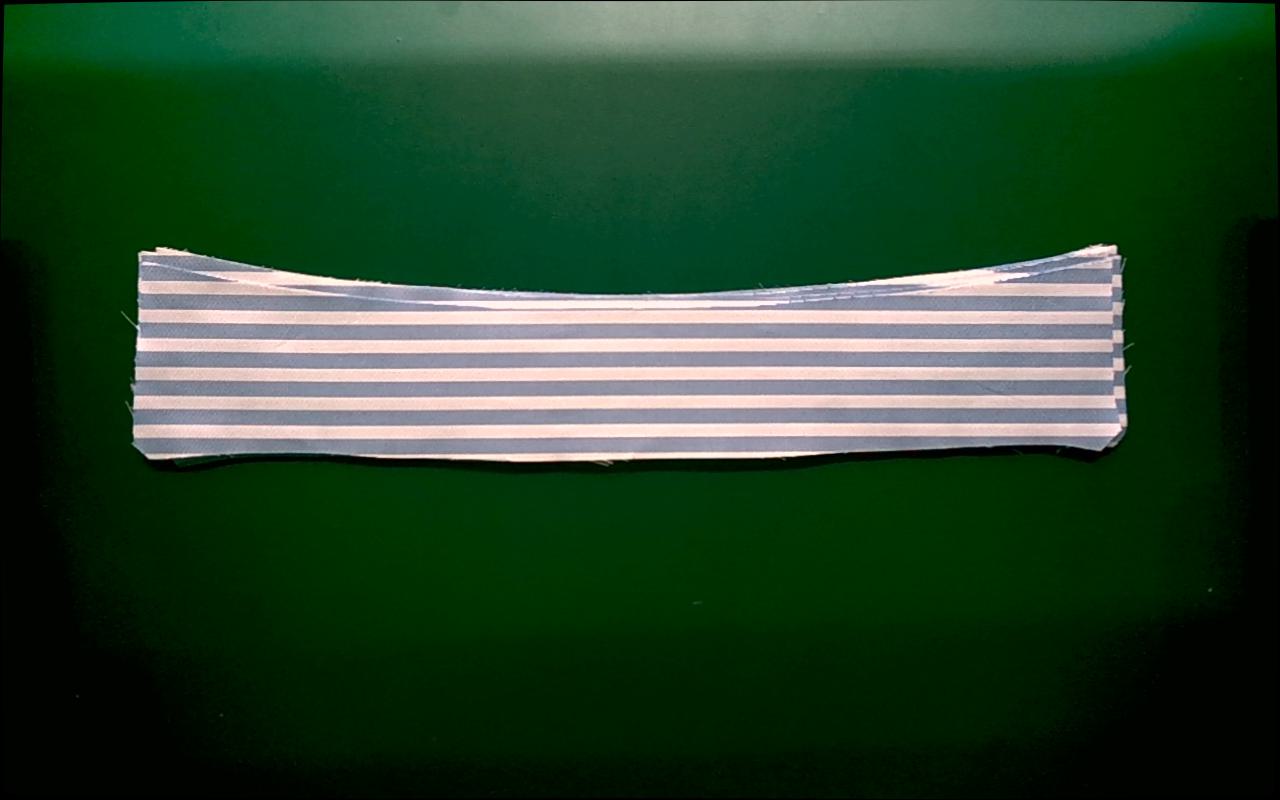}
        \caption*{Input image}
    \end{minipage}%
    \hspace{0.001\textwidth}
    \begin{minipage}[t]{0.154\textwidth}
        \centering
        \includegraphics[width=0.95\textwidth]{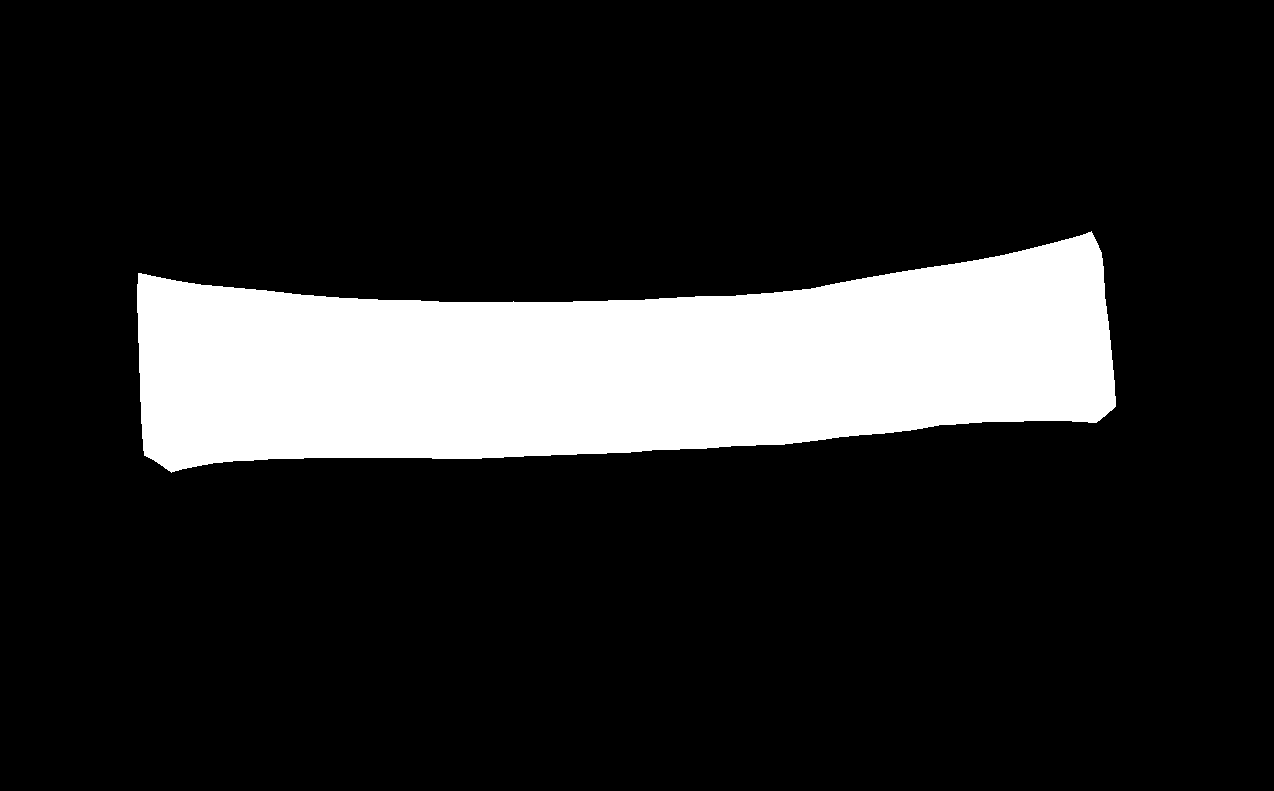}\par\vspace{0.07\textwidth}
        \includegraphics[width=0.95\textwidth]{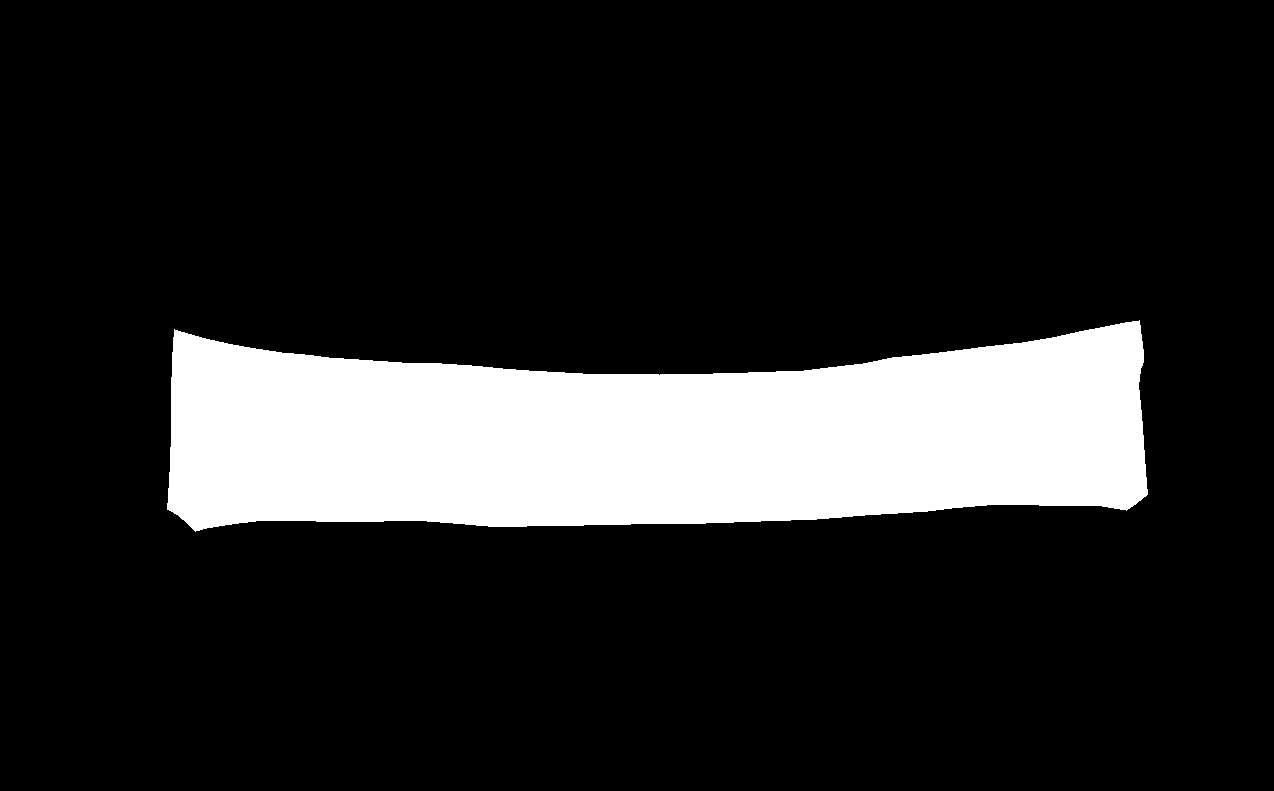}\par\vspace{0.07\textwidth}
        \includegraphics[width=0.95\textwidth]{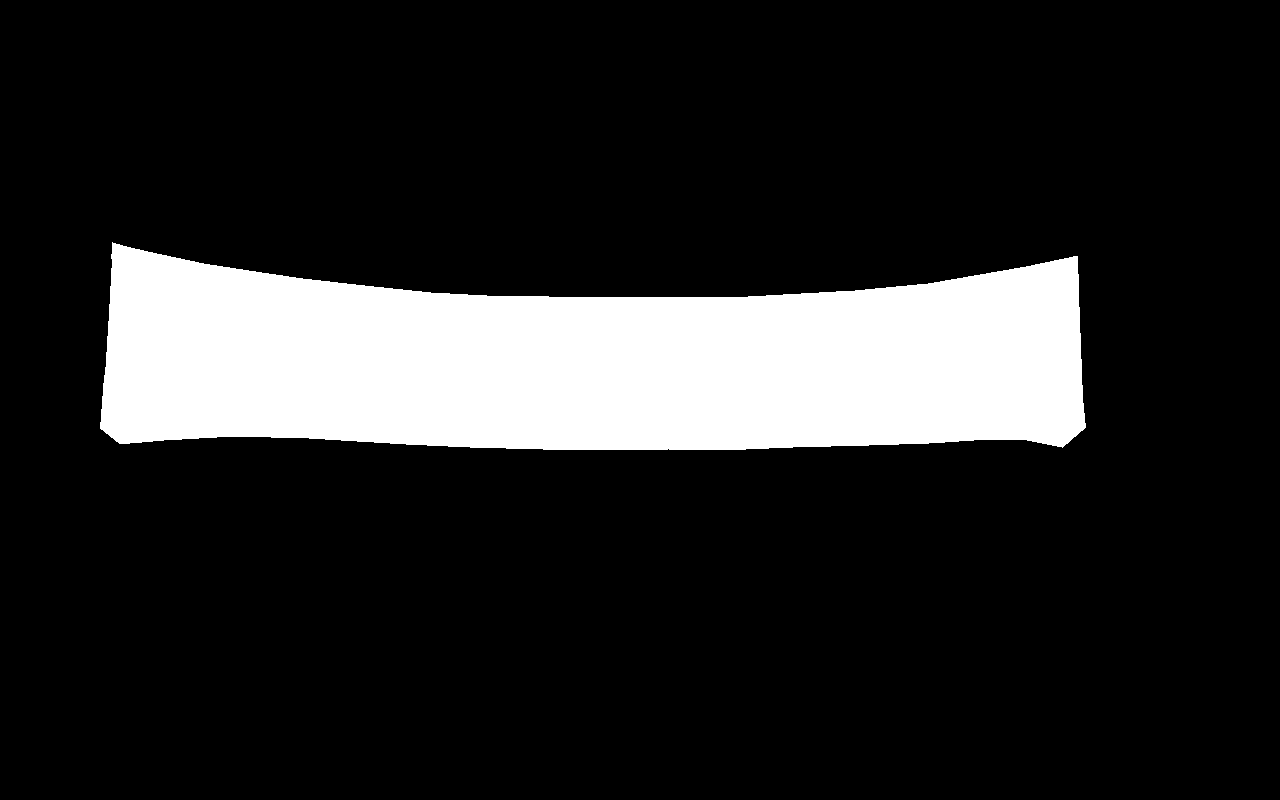}\par\vspace{0.07\textwidth}
        \includegraphics[width=0.95\textwidth]{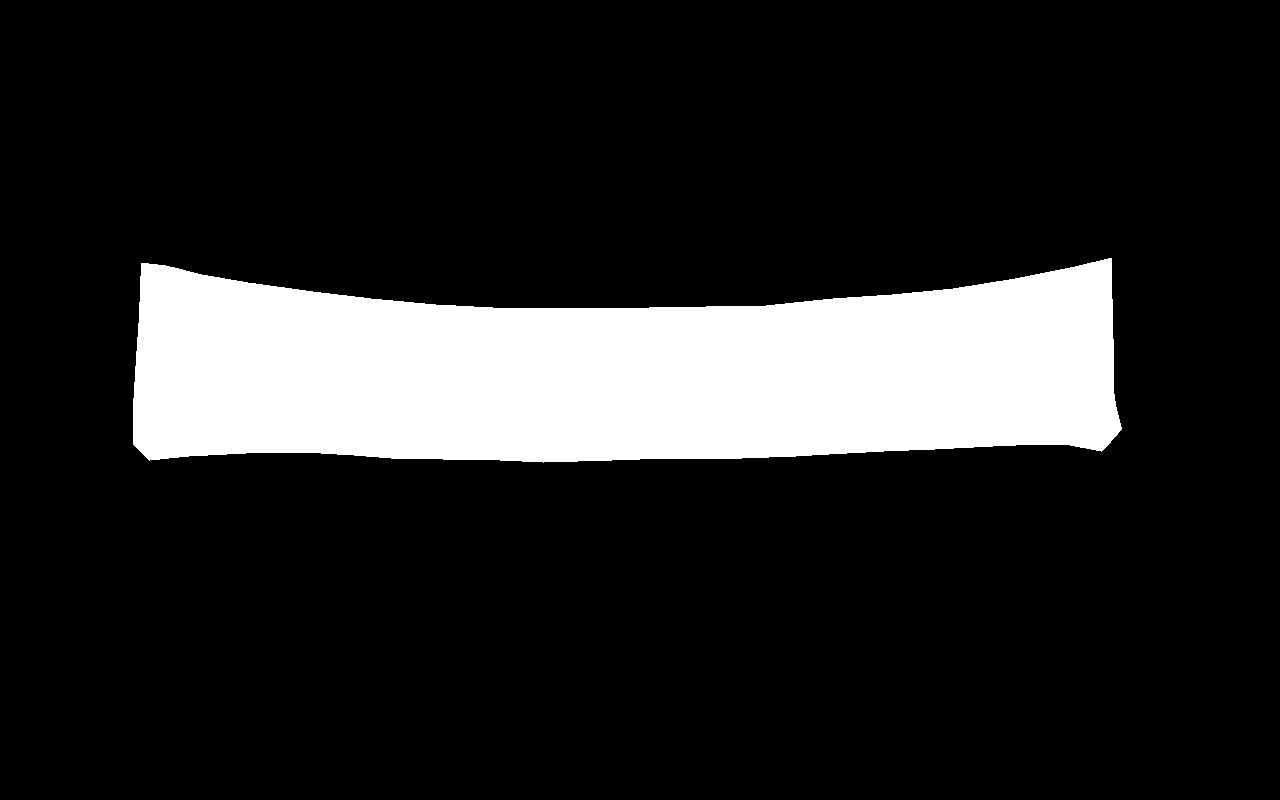}
        \caption*{Ground-truth}
    \end{minipage}%
    \hspace{0.001\textwidth}
    \begin{minipage}[t]{0.154\textwidth}
        \centering
        \includegraphics[width=0.95\textwidth]{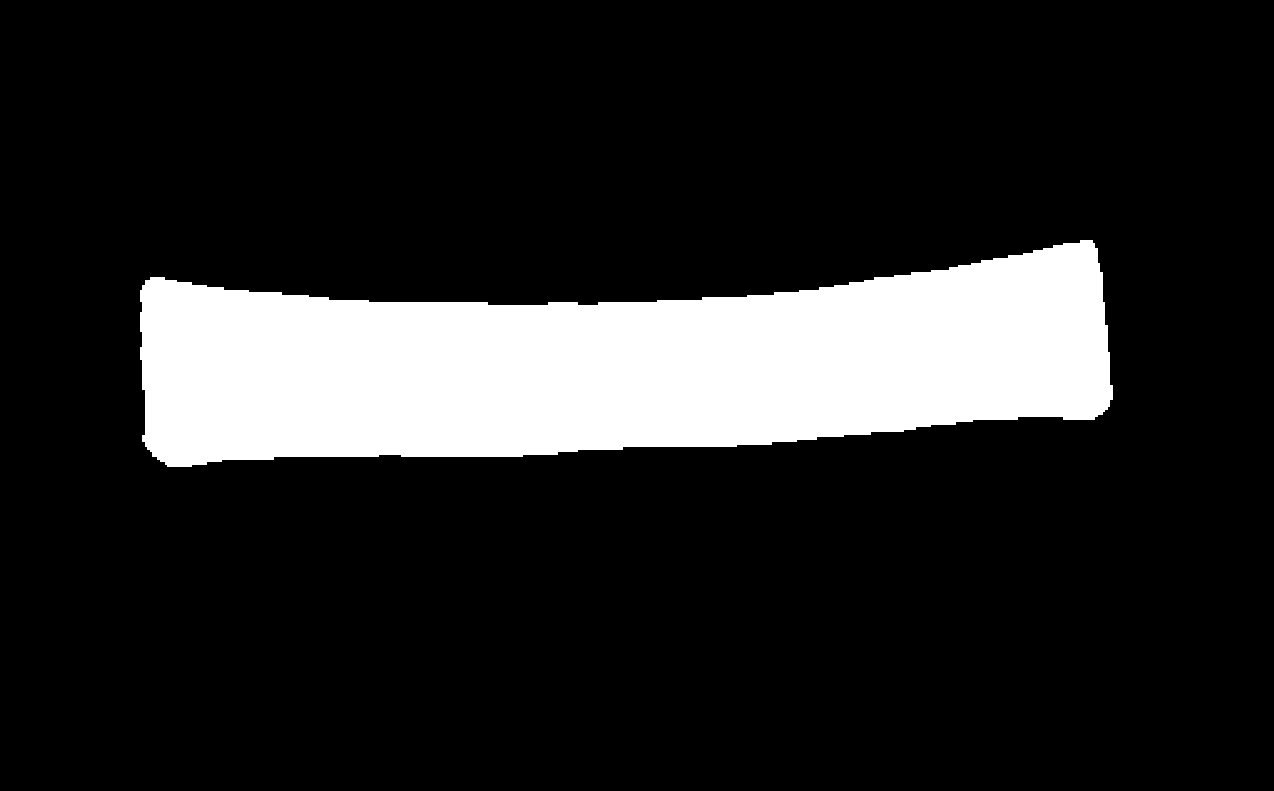}\par\vspace{0.07\textwidth}
        \includegraphics[width=0.95\textwidth]{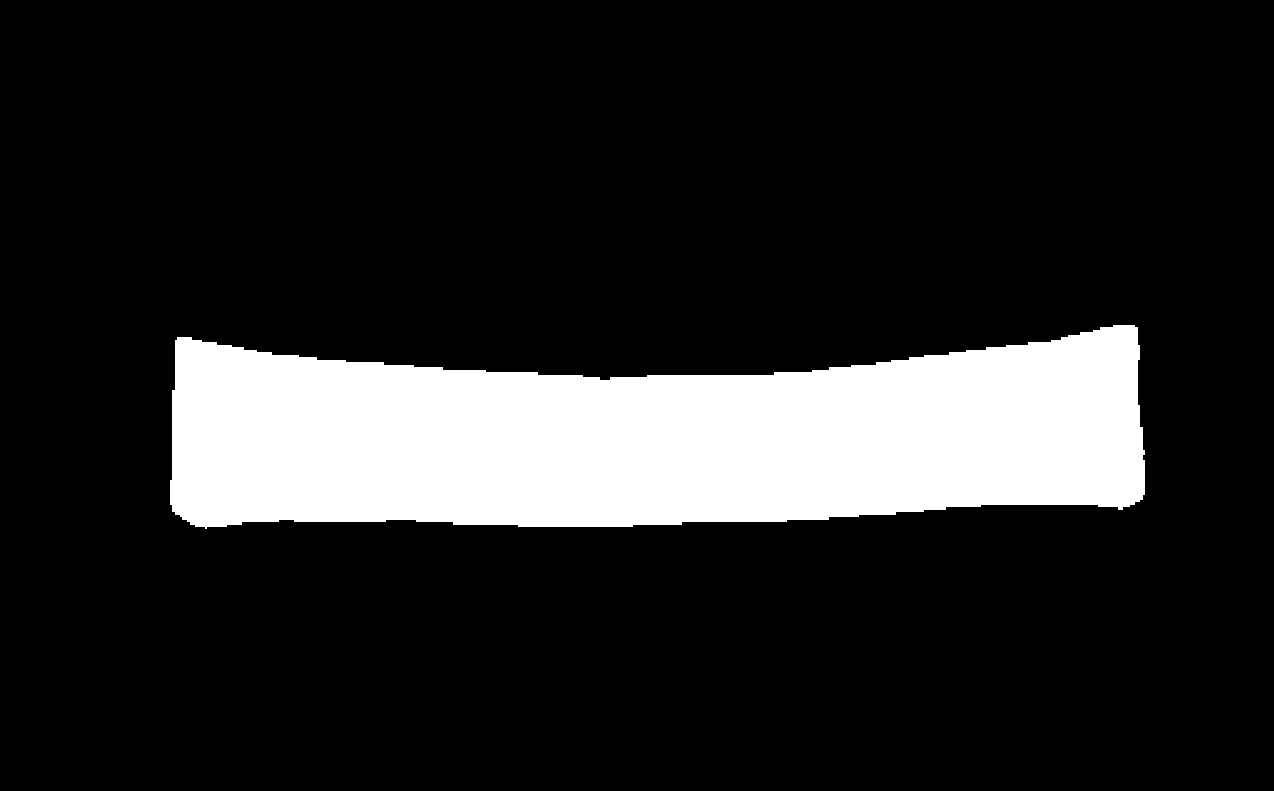}\par\vspace{0.07\textwidth}
        \includegraphics[width=0.95\textwidth]{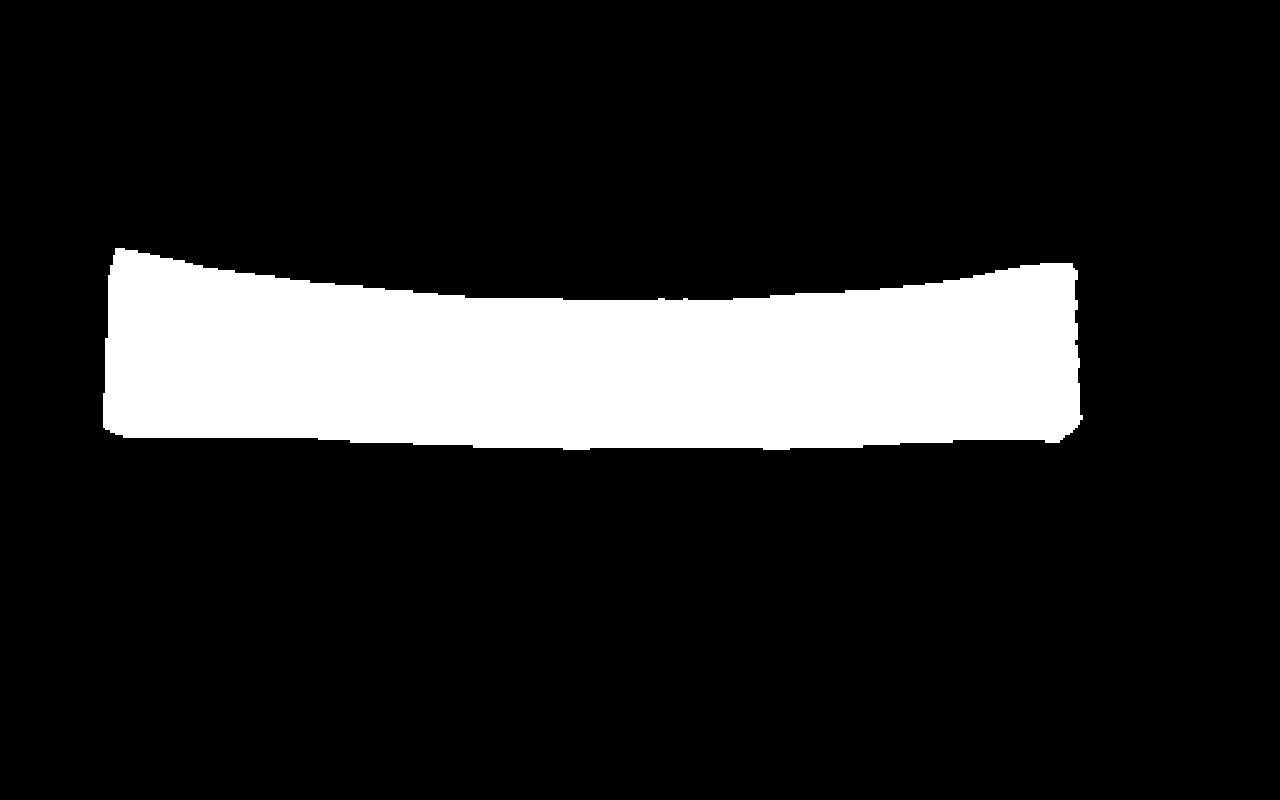}\par\vspace{0.07\textwidth}
        \includegraphics[width=0.95\textwidth]{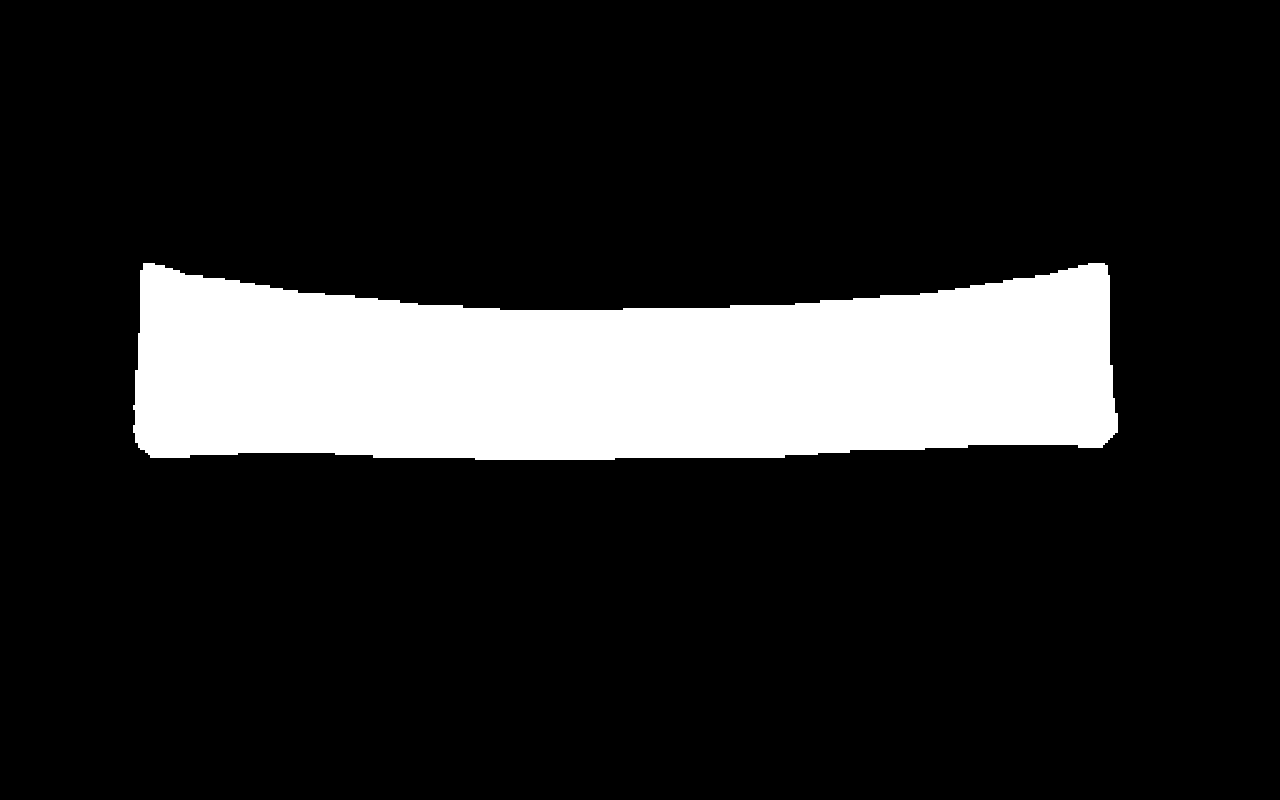}
        \caption*{Prediction}
    \end{minipage}
    \caption{Qualitative segmentation results. Quantitative comparison is included in Table~\ref{tab:ablation}.}
    \label{fig:qualitative}
\end{figure}


\textbf{Dataset description:} To evaluate the effectiveness of the proposed network, experiments were conducted on a real collar fabric dataset comprising 235 images with corresponding pixel-level annotations. Each image is accompanied by a semantic label mask, a ground-truth edge mask, and an ideal shape label derived from the corresponding CAD model.


\textbf{Implementation details:} The proposed network is implemented using PyTorch. The Adam optimizer is employed, using an initial learning rate of $1 \times 10^{-4}$ and a batch size of 8. Training is conducted over 300 epochs, with early stopping applied based on validation loss to prevent overfitting. All experiments are performed on a workstation equipped with an NVIDIA RTX 3060 GPU (12GB VRAM), an Intel Core i9-10900F CPU, and 16 GB of RAM. Model training and evaluation are conducted under CUDA 11.7 with cuDNN acceleration enabled.


\textbf{Evaluation metrics:} To comprehensively assess the segmentation performance, the following standard metrics were utilized

\begin{enumerate}
    \item \textbf{Intersection over Union (IoU):} Measures the overlap between the predicted and ground truth regions, and is defined as
    \begin{equation}
        \mathrm{IoU} = \frac{TP}{TP + FP + FN},
    \end{equation}
    where $TP$, $FP$, and $FN$ denote the numbers of true positive, false positive, and false negative pixels, respectively.
    
    \item \textbf{Pixel Accuracy (PA):} Computes the proportion of correctly predicted pixels:
    \begin{equation}
        \mathrm{PA} = \frac{TP + TN}{TP + FP + FN + TN},
    \end{equation}
    where $TN$ denotes the number of true negative pixels.
    
    \item \textbf{Edge Mean Squared Error (EMSE) :} 
    This and the following metrics evaluate the spatial accuracy of the predicted segmentation boundaries. 
    Given the set of predicted edge points $\mathcal{P} = \{\mathbf{p}_i\}_{i=1}^N$ and the set of edge ground truth points $\mathcal{G} = \{\mathbf{g}_j\}_{j=1}^M$, 
    the EMSE is defined as the average squared Euclidean distance between each predicted edge point and its nearest ground truth edge point, and between each ground truth edge point and its nearest predicted edge point, averaged over both directions, and given by
    \begin{align}
    \mathrm{EMSE} &= \frac{1}{2} \left( \mathrm{MSE}_{\mathcal{P}\rightarrow\mathcal{G}} + \mathrm{MSE}_{\mathcal{G}\rightarrow\mathcal{P}} \right),
    \end{align}
    where
    \begin{align}
    \mathrm{MSE}_{\mathcal{P}\rightarrow\mathcal{G}} &= \frac{1}{N}\sum_{i=1}^{N} \left( \min_{\mathbf{g}_j \in \mathcal{G}} \|\mathbf{p}_i - \mathbf{g}_j\|_2 \right)^2 \nonumber, \\
    \mathrm{MSE}_{\mathcal{G}\rightarrow\mathcal{P}} &= \frac{1}{M}\sum_{j=1}^{M} \left( \min_{\mathbf{p}_i \in \mathcal{P}} \|\mathbf{g}_j - \mathbf{p}_i\|_2 \right)^2\nonumber. 
    \end{align}
   \item \textbf{Edge Root Mean Squared Error (ERMSE):} The ERMSE is then defined as
    \begin{equation}
        \mathrm{ERMSE} = \frac{1}{2} \left( \sqrt{\mathrm{MSE}_{\mathcal{P}\rightarrow\mathcal{G}}} + \sqrt{\mathrm{MSE}_{\mathcal{G}\rightarrow\mathcal{P}}} \right),
    \end{equation}
    where $\|\cdot\|_2$ denotes the Euclidean distance between two points.
\end{enumerate}


\textbf{Ablation study:} To analyze the contribution of each architectural component, ablation studies were conducted. Specifically, the following network architectures were compared
\begin{enumerate}
    \item \textbf{Baseline:} Standard segmentation network without edge- or shape-aware branches.
    \item \textbf{Baseline + edge-aware branch:} Segmentation network with added edge-aware branch only.
    \item \textbf{Baseline + edge-aware branch + shape-aware branch (Proposed architecture)}: Full model with both edge-aware branch and shape-aware branch.
\end{enumerate}

Table~\ref{tab:ablation} summarizes the quantitative results. The addition of the edge-aware branch improves both IoU and PA, while also reducing the ERMSE. This demonstrates that the edge-aware branch helps the network produce more accurate and sharper segmentation masks. Adding the shape-aware branch after the edge-aware branch further improves all performance metrics. In particular, the shape-aware branch enhances the network's ability to capture the global structure of the fabric, resulting in higher IoU and PA, as well as the lowest ERMSE, which indicates improved alignment between the predicted and ideal shapes. These results confirm that both the edge-aware and shape-aware branches are beneficial and complementary for achieving accurate segmentation.
\begin{table}[h]
    \centering
    \caption{Ablation study of architectural components.}
    \label{tab:ablation}
    \resizebox{\linewidth}{!}{%
    \begin{tabular}{lccc}
        \toprule
        \textbf{Model} & \textbf{IoU $\uparrow$ (\%)} & \textbf{PA $\uparrow$ (\%)} & \textbf{ERMSE $\downarrow$ (pixel)} \\
        \midrule
        Baseline       &  93.25 & 95.87 & 5.03  \\
        Baseline + EA      &  96.24 & 96.72 & 3.19  \\
        Baseline + EA + SA (Proposed architecture) &  \textbf{96.80} & \textbf{97.50} & \textbf{2.58} \\
        \bottomrule
    \end{tabular}
    }
    \vspace{1pt}
    \\
    {\footnotesize \textbf{Note:} EA: edge-aware branch; SA: shape-aware branch.}
\end{table}

%

\textbf{Qualitative results:} Figure~\ref{fig:qualitative} presents qualitative segmentation results. As shown, the proposed architecture accurately delineates fabric boundaries and preserves the overall shape, even in challenging regions with ambiguous edges and complex contours. These qualitative results demonstrate that our method produces precise boundaries while maintaining the global integrity of the fabric shape.

\section{Conclusion}

This paper presents a novel segmentation training architecture designed specifically for the challenging task of top-layer fabric segmentation in stacked scenarios. By integrating an edge-aware branch to refine boundary localization and a shape-aware branch to align predicted masks with Computer Aided Design (CAD) models, the proposed method effectively supervises and tunes the backbone network to address challenges from subtle fabric boundaries and high inter-layer similarity. Experiments on real-world fabric datasets demonstrate that both supervisory branches enhance segmentation accuracy, especially at fabric edges. The architecture also shows robustness to limited training data and outperforms strong baselines.

%
\textbf{Future work:} The proposed top-layer fabric segmentation method will be implemented in our pick-and-place system~\cite{kobayashi2025rollup}, and will demonstrate its versatility and practical applicability as future work.


\section*{Acknowledgment}

This work was supported in part by the Innovation and Technology Commission of the HKSAR Government under the InnoHK initiative. The research described in this paper was conducted in part at the JC STEM Lab of Robotics for Soft Materials, funded by The Hong Kong Jockey Club Charities Trust.

\bibliographystyle{IEEEtran}
\bibliography{reference}

\end{document}